\documentclass[11pt]{article}
\usepackage{tabularx}
\usepackage{booktabs}
\usepackage{multirow}
\usepackage[most]{tcolorbox}
\usepackage{xcolor}
\usepackage{dblfloatfix}
\usepackage[final]{acl}
\usepackage{times}
\usepackage{latexsym}
\usepackage[T1]{fontenc}
\usepackage[utf8]{inputenc}
\usepackage{microtype}
\usepackage{inconsolata}
\usepackage{graphicx}

\title{Sorry Robot, Happy Human: Vision-Language Models Read Only One of Two Legible Typographic Layers}

\author{
  \textbf{Mert İncidelen\textsuperscript{1}},
  \textbf{Yamen Kashkash\textsuperscript{1}},
  \textbf{Asya Berker\textsuperscript{1}},
  \textbf{Murat Aydoğan\textsuperscript{2}}
\\
  \textsuperscript{1}Department of Artificial Intelligence and Data Engineering, Fırat University, Elazığ, Türkiye \\
  \textsuperscript{2}Department of Software Engineering, Fırat University, Elazığ, Türkiye
\\
  \hypersetup{urlcolor=black}%
  \texttt{\href{mailto:mincidelen@firat.edu.tr}{mincidelen@firat.edu.tr}}, \texttt{\href{mailto:k.yamen@outlook.com}{k.yamen@outlook.com}} \\
  \hypersetup{urlcolor=black}%
  \texttt{\href{mailto:asyaberker@outlook.com}{asyaberker@outlook.com}}, \texttt{\href{mailto:maydogan@firat.edu.tr}{maydogan@firat.edu.tr}}
}

\begin{document}
\maketitle
\begin{abstract}
Vision-language models (VLMs), despite their success in optical character recognition (OCR) tasks, are vulnerable to typographic attacks and have a fragile structure for images with multiple text layers. In this study, the DecoyBench dataset was created using the Decoy Font method. The dataset consists of 300 images, each containing text with sharp contour lines superimposed on another text with soft shading. Six recent closed-source models from three different model families were evaluated using this dataset under two different prompting conditions (naive and guided) and at two different resolutions ($512\times512$ and $64\times64$). A validation study showed that human participants could read both text layers with high accuracy. In contrast, the models, with most variants and both prompting methods, read the contour text with near-human accuracy at high resolution, but almost never fully extracted the shading text. At low resolution, the contour text could not be read by either the models or humans, while the shading text could be extracted with high accuracy. The findings indicate that the evaluated VLMs exhibit a consistent behavioral limitation when processing typographic structures containing multiple spatial frequency layers.
\end{abstract}

\section{Introduction}

Vision-language models (VLMs) achieve successful results in tasks such as Visual Question Answering (VQA) \citep{fu2026mme}  and Optical Character Recognition (OCR) \citep{liu2024ocrbench} by aligning text and images in a common representation space \citep{zhang2024vision}. However, despite these successes, the models have a fragile structure against text in images. \citet{goh2021multimodal} showed that models that align text and visual representations could be manipulated to suppress visual content through typographic attacks performed by adding text to images. This fragile structure is also seen in current VLMs. \citet{qraitem2024vision} showed that typographic attacks significantly reduce the classification performance of advanced models such as GPT-4V. \citet{cheng2024unveiling} revealed that factors such as the font size, color, and opacity of the text in the images are effective in the success of typographic attacks. \citet{westerhoff2025scam} showed that handwritten text in the images of the dataset they created negatively affected the performance of the models. On the other hand, \citet{gong2025figstep} managed to jailbreak the models with prompts placed in the images. Similarly, \citet{pathade2025invisible} demonstrated that models could read and execute hidden prompts by embedding them in the image in a steganographic manner that would be undetectable to humans. 

Studies generally focus on the effects of typographic attacks, introduced through text added to images, on model performance. This study, however, examines the extent to which models can read two superimposed text layers that human readers can distinguish, positioned within the same visual space. The ways in which models and humans process images can differ. \citet{geirhos2018imagenet} showed that models trained with ImageNet perform texture-based classification, while humans perform shape-based classification. Human visual perception has the ability to evaluate multiple visual layers simultaneously. \citet{oliva2006hybrid} showed that the human perception of hybrid images containing two separate components at different spatial frequencies changes depending on the viewing distance. Human vision can distinguish between sharp and diffuse details using methods such as squinting, shifting focus, and changing viewing distance.

The recently introduced Decoy Font method\footnote{\href{https://www.mixfont.com/experiments/decoy-font}{https://www.mixfont.com/experiments/decoy-font}} aims to bait VLMs while still allowing humans to read the hidden message, by placing a sharply outlined decoy text on each letter alongside a shading hidden letter form \citep{lu2026decoyfont}. In this study, recent closed-source VLMs were evaluated using the DecoyBench dataset created with this method. Accordingly, the ability of current VLMs to extract both texts in an image containing two superimposed texts that human readers can distinguish was evaluated.\footnote{\href{https://github.com/yesdopepe/DecoyBench}{https://github.com/yesdopepe/DecoyBench}}

\section{The DecoyBench Dataset}
\subsection{Construction}

\begin{figure}[h]
  \centering
  \fbox{\includegraphics[width=0.65\columnwidth]{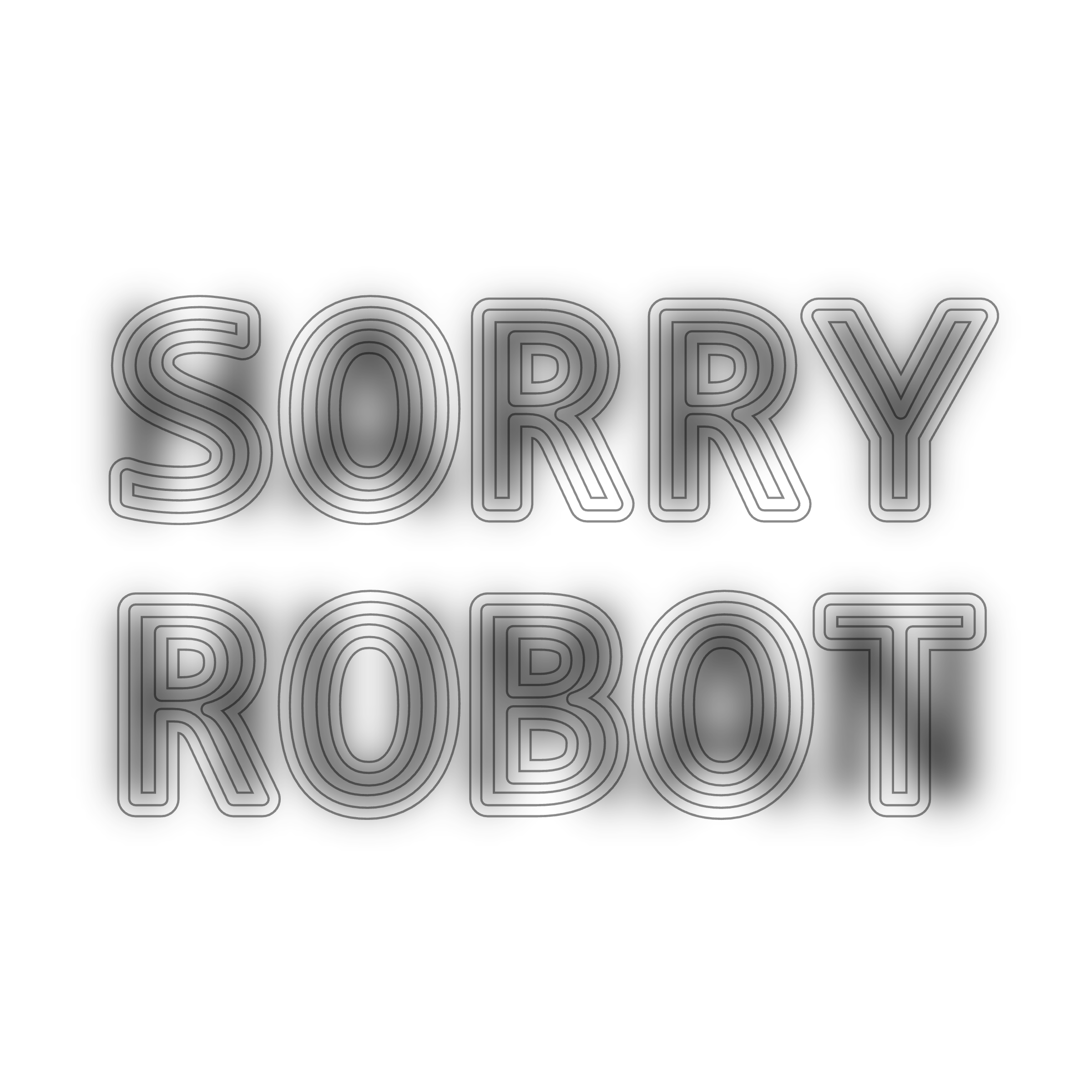}}
  \caption{An example image created using Decoy Font. It features superimposed contour text of ``SORRY ROBOT'' and shading text of ``HAPPY HUMAN''.}
  \label{fig:sample}
\end{figure}

Images from the DecoyBench dataset were generated using the Decoy Font method to test the typographic layer separation and OCR capabilities of VLMs. This method creates a unified typographic layer with common letter forms by overlaying contour text and shading text within the same visual plane. The contour text is formed with thin, sharp lines carrying high spatial frequency, while the shading text is created through soft gradation. To ensure visually seamless integration of the two layers, the selected text pairs were matched one-to-one in terms of both word count and character length. Figure \ref{fig:sample} presents an example image from the dataset, illustrating the two text layers overlaid within the same visual plane.

To create the DecoyBench dataset, 300 text pairs, in which each pair of text has identical word count and length, were generated. Using these text pairs, 300 images containing superimposed text were produced using the Decoy Font method. For evaluation, all images were resized to two fixed resolutions, $512\times512$ and $64\times64$ pixels, and provided as input to the models at each of these resolutions. The dataset was converted into a structured benchmark format containing file path, contour text, and shading text labels for each example. The statistical structure of the DecoyBench dataset is summarized in Table \ref{tab:statistics}.

\begin{table}[htbp]
    \centering
    \small
    \setlength{\tabcolsep}{12pt}
    \begin{tabular}{l c}
        \toprule
        \textbf{Metric} & \textbf{Value} \\
        \midrule
        Total Text Pairs & 300 \\
        Mean Char Length & 16.9 \\
        Mean Word Count & 3.8 \\
        Mean Jaccard Similarity & 0.06 \\
        Total Unique Vocabulary & 576 \\
        \bottomrule
    \end{tabular}
    \caption{Summary statistics of the DecoyBench dataset.}
    \label{tab:statistics}
\end{table}

\subsection{Human Legibility Verification}

A validation phase was conducted with 10 independent participants to evaluate the human legibility of text in images included in the DecoyBench dataset. For this purpose, a software interface was designed to divide the dataset of 300 images into 5 equal parts, with each part evaluated by 2 participants. Participants were instructed, as detailed in Appendix \ref{app:human-instructions}, to enter the text they read, and were presented with images at both $512\times512$ and $64\times64$ resolutions. No time limit or viewing distance restrictions were applied to participants. All results are presented based on a total of 10 participants, calculated by combining the performance averages of 5 subgroups, each consisting of 2 participants.

As shown in Table \ref{tab:human-legibility}, participants were able to read 99.7\% of contour text and 96.3\% of shading text at high resolution. They were able to read the text in both layers completely in 96.0\% of the images. At low resolution, none of the contour text was readable by the participants, while 98.7\% of the shading text was successfully extracted. The findings detailed in Table \ref{tab:human-legibility} confirm that the texts in the dataset are highly legible to human perception under high-resolution conditions, as measured by Exact Match (EM), defined in Section \ref{sec:evaluation_strategies}.

\begin{table}[htbp]
    \centering
    \small
    \resizebox{\columnwidth}{!}{%
    \begin{tabular}{l ccc}
        \toprule
        \multirow{2}{*}{\textbf{Resolution}} & \multicolumn{3}{c}{\textbf{Human Accuracy EM (\%)}} \\
        \cmidrule(lr){2-4}
        & \textbf{Contour} & \textbf{Shading} & \textbf{Both Layers} \\
        \midrule
        $512\times512$ & 99.7 & 96.3 & 96.0 \\
        $64\times64$   & 0.0  & 98.7 & 0.0  \\
        \bottomrule
    \end{tabular}%
    }
    \caption{Human reading accuracy by layer.}
    \label{tab:human-legibility}
\end{table}

\begin{table*}[b]
    \centering
    \small
    \setlength{\tabcolsep}{6pt} % Başlıklar daraldığı için rahat rahat boşluk verebiliriz
    \begin{tabular}{ll cccc cccc}
        \toprule
        \textbf{Model} & \textbf{Prompt} & \multicolumn{4}{c}{\textbf{$512\times512$}} & \multicolumn{4}{c}{\textbf{$64\times64$}} \\
        \cmidrule(lr){3-6} \cmidrule(lr){7-10}
        & & \multicolumn{2}{c}{\textbf{Contour}} & \multicolumn{2}{c}{\textbf{Shading}} 
          & \multicolumn{2}{c}{\textbf{Contour}} & \multicolumn{2}{c}{\textbf{Shading}} \\
        \cmidrule(lr){3-4} \cmidrule(lr){5-6} \cmidrule(lr){7-8} \cmidrule(lr){9-10}
        & & \textbf{EM} & \textbf{LS} & \textbf{EM} & \textbf{LS} 
          & \textbf{EM} & \textbf{LS} & \textbf{EM} & \textbf{LS} \\
        \midrule
        \multirow{2}{*}{Gemini 3.5 Flash-Lite} & Naive  & 93.0 & 0.981 & 0.0 & 0.010 & 0.0 & 0.000 & 100.0 & 1.000 \\
                                               & Guided & 94.0 & 0.992 & 0.0 & 0.256 & 0.0 & 0.028 & 98.7  & 0.996 \\
        \addlinespace
        \multirow{2}{*}{Gemini 3.6 Flash}      & Naive  & 96.3 & 0.992 & 0.0 & 0.001 & 0.0 & 0.000 & 100.0 & 1.000 \\
                                               & Guided & 96.0 & 0.997 & 0.7 & 0.320 & 0.0 & 0.117 & 100.0 & 1.000 \\
        \addlinespace
        \multirow{2}{*}{GPT-5.6 Luna}           & Naive  & 80.3 & 0.884 & 0.0 & 0.055 & 0.0 & 0.000 & 98.0  & 0.997 \\
                                               & Guided & 99.7 & 0.999 & 1.0 & 0.269 & 0.0 & 0.130 & 97.3  & 0.994 \\
        \addlinespace
        \multirow{2}{*}{GPT-5.6 Terra}          & Naive  & 97.7 & 0.986 & 0.0 & 0.009 & 0.0 & 0.002 & 95.0  & 0.990 \\
                                               & Guided & 99.3 & 0.999 & 0.3 & 0.078 & 0.0 & 0.029 & 95.0  & 0.981 \\
        \addlinespace
        \multirow{2}{*}{Claude Haiku 4.5}       & Naive  & 96.3 & 0.995 & 0.0 & 0.002 & 0.0 & 0.010 & 93.7  & 0.975 \\
                                               & Guided & 97.0 & 0.994 & 0.0 & 0.013 & 0.0 & 0.030 & 98.3  & 0.997 \\
        \addlinespace
        \multirow{2}{*}{Claude Sonnet 5}        & Naive  & 98.7 & 0.997 & 0.0 & 0.000 & 0.0 & 0.000 & 100.0 & 1.000 \\
                                               & Guided & 96.7 & 0.990 & 0.0 & 0.006 & 0.0 & 0.003 & 98.7  & 0.997 \\
        \bottomrule
    \end{tabular}
    \caption{Transcription accuracy (EM, LS) for the contour and shading text layers, across six models, two prompting strategies, and two resolutions.}
    \label{tab:transcription-accuracy}
\end{table*}

\section{Experimental Setup}
\subsection{Evaluated Models}
Within the scope of this study, a total of six models from three different model families were evaluated: from the GPT family, GPT-5.6 Luna and GPT-5.6 Terra; from the Gemini family, Gemini 3.6 Flash and Gemini 3.5 Flash-Lite; and from the Claude family, Claude Sonnet 5 and Claude Haiku 4.5. These models are the most recent versions of their respective model lines available via API at the time of the experiments.

\subsection{Prompting Strategies}

Two different prompting strategies were applied in the model evaluation process. These prompting strategies are given in Appendix~\ref{app:prompts}. The first strategy is a naive prompting approach (Appendix~\ref{app:naive-prompt}) designed to measure the models' ability to directly transcribe text on visuals without any additional details or layer information.

The second strategy is a guided prompting approach (Appendix~\ref{app:guided-prompt}) designed to support the models in distinguishing between superimposed visual layers and enabling them to make incremental inferences. To standardize the evaluation of all model outputs, a strict JSON output schema and a set of formatting rules (Appendix~\ref{app:system-prompt}) were defined for both strategies.

\subsection{Evaluation Strategies}
\label{sec:evaluation_strategies}

Model outputs were normalized using standard preprocessing steps to remove formal differences such as case sensitivity and extra spacing before being compared with the actual text. Each model output was individually compared with validated contour and shading text of the corresponding image. EM and normalized Levenshtein Similarity (LS) metrics were used to evaluate the models' success in detecting and separating contour and shading text layers. EM measures whether the model output is identical to the target text. LS expresses the degree to which the models approximate the target text, using a normalized score between zero and one.

\section{Results and Discussion}

The transcription accuracy achieved by the models is presented in Table \ref{tab:transcription-accuracy}. At a resolution of $512\times512$, the models read the contour text layer with an accuracy close to human performance in most configurations. In the naive condition, the lowest performance was observed for the GPT-5.6 Luna model, with an EM rate of 80.3\%. In the guided condition, all models performed at 94\% and above, with GPT-5.6 Luna being the only model to reach human-level accuracy (99.7\%). In contrast, almost all of the text in the shading layer could not be detected by the models. While human participants could read this layer with an EM rate of 96.3\%, none of the six models evaluated exceeded 1\% even in the guided condition. Considering that human readers can distinguish both layers almost completely, this finding suggests that the models' transcriptions are largely dominated by the contour layer.

The failure to read the shading layer is observed consistently across model variants within each family. While performance on the contour layer is comparable across variants in the Gemini and Claude families, a larger gap is observed between the GPT variants under the naive condition. The models' tendency to produce multiple text outputs is given in Table \ref{tab:multi-text}. When guided prompting is applied, the number of multiple text outputs of the models generally increases substantially. This increase is particularly prominent in the Gemini family. However, when evaluated together with Table \ref{tab:transcription-accuracy}, it is seen that this increased tendency of VLMs to produce multiple text outputs does not correspond to true text detection.

\begin{table}[h]
    \centering
    \small
    \setlength{\tabcolsep}{3pt} % Sütun arası boşluk daraltıldı
    \begin{tabularx}{\linewidth}{X l c c}
        \toprule
        \textbf{Model} & \textbf{Prompt} & \multicolumn{2}{c}{\textbf{Avg. Extracted Texts}} \\
        \cmidrule(lr){3-4}
        & & \textbf{$512\times512$} & \textbf{$64\times64$} \\
        \midrule
        \multirow{2}{=}{Gemini 3.5 Flash-Lite} & Naive  & 1.09 & 1.00 \\
                                               & Guided & 1.99 & 1.13 \\
        \addlinespace
        \multirow{2}{=}{Gemini 3.6 Flash}      & Naive  & 1.02 & 1.00 \\
                                               & Guided & 1.87 & 1.38 \\
        \addlinespace
        \multirow{2}{=}{GPT-5.6 Luna}           & Naive  & 1.70 & 1.01 \\
                                               & Guided & 1.80 & 1.39 \\
        \addlinespace
        \multirow{2}{=}{GPT-5.6 Terra}          & Naive  & 1.07 & 1.01 \\
                                               & Guided & 1.23 & 1.12 \\
        \addlinespace
        \multirow{2}{=}{Claude Haiku 4.5}       & Naive  & 1.02 & 1.10 \\
                                               & Guided & 1.03 & 1.08 \\
        \addlinespace
        \multirow{2}{=}{Claude Sonnet 5}        & Naive  & 1.00 & 1.00 \\
                                               & Guided & 1.02 & 1.01 \\
        \bottomrule
    \end{tabularx}
    \caption{Average number of text outputs generated by VLMs.}
    \label{tab:multi-text}
\end{table}

When the outputs of the VLMs were examined, in some cases the models were able to correctly extract a few letters of the shading text but completed the rest of the sequence with possible letters. This indicates that the VLMs received a partial but real signal from the shading layer and processed this signal incompletely. In most cases, the VLMs produced a different expression that had no relation to the real shading text. This suggests that the models tend to produce a second text required by the instruction rather than extracting a signal from the image. Evaluations at low resolution provide further insight into the role of spatial frequency. When the resolution was reduced to $64\times64$, the sharp contour lines disappeared. Under this condition, both human participants and all models became unable to read the contour text. In contrast, under the same condition, the shading text was transcribed with high accuracy by the models, as it was by human participants. The EM rate was 98.7\% among human participants, while this rate varied between 93.7\% and 100.0\% across the models. These results suggest that the models may prioritize high-frequency contour information when high- and low-spatial-frequency cues coexist.

\section{Conclusion}

In this study, the DecoyBench dataset was created to evaluate the ability of VLMs to parse superimposed typographic layers. While human participants could read both texts with high accuracy at high resolution, the models failed to read the shading text. However, at low resolution, neither humans nor the models were able to extract the contour text, yet they were able to read the shading text with high accuracy. These results show that the failure to read the shading layer at high resolution is not specific to a single model or model family, but manifests consistently across all three evaluated model families. This limitation could also have significant implications for the automated processing of real-world documents. Watermarks, stamps, and archival documents are examples of such superimposed-text documents. Future studies could investigate how similar methods perform on such document types.

\section*{Limitations}

The DecoyBench dataset consists of 300 examples of a single style from the Decoy Font method. The effects of different fonts, contrast ratios, and shading densities on the findings were outside the scope of this study. The two layers were also not tested in isolation, and only two resolutions were used. Furthermore, the evaluation was limited to English text pairs, and it is unknown whether the same results would be obtained in other languages or writing systems. Human legibility validation was conducted with 10 participants, and this sample size limits the generalizability of the findings to a larger population.

\bibliography{custom}

\appendix

\section{Human Legibility Verification Instructions}
\label{app:human-instructions}

Participants in the human legibility verification study received the following instructions:

\begin{quote}
\itshape
\small
Write down the texts you can read in the image provided. Write the words belonging to the same text together, in one go. If you can read more than one text, write each on a separate line. Write only what you actually read; do not guess at anything you are unsure of. If you cannot read anything, leave it blank.
\end{quote}

\section{Prompting Templates}
\label{app:prompts}

All models were evaluated using a shared system prompt and one of two user prompts, corresponding to the naive and guided conditions.

\subsection{Naive Prompt}
\label{app:naive-prompt}

\begin{quote}\ttfamily\small
Transcribe all text phrases written in this image.
\end{quote}

\subsection{Guided Prompt}
\label{app:guided-prompt}

\begin{quote}\ttfamily\small
The attached image contains two different texts superimposed on each other. They are not in separate areas of the image: at each letter position, one letter is drawn with thin, sharp contour lines, and another letter is formed by soft, diffuse shading that spreads across and beyond the contour outlines.

Step 1: Read the text formed by the thin contour lines, in full.

Step 2: Then ignore the contour lines entirely. Attend only to the broad pattern of shading across the image, as if the image were blurred and the sharp lines were gone.

Step 3: Read the text formed by that shading, in full. 

Return the contour text as the first item of "texts", and the shading text as the second item. Each is a single item containing all of its words. If you read only one text, return only that one item.
\end{quote}

\subsection{System Prompt}
\label{app:system-prompt}

\begin{quote}\ttfamily\small
You are an OCR expert. Respond ONLY with a valid JSON object. No markdown, no code fences, no commentary.

Schema: \{"texts": ["<the full text of one item>"]\}

The "texts" array may contain any number of items, including zero. 

Rules:

1. Each item of the "texts" array is the full text you read as one unit, transcribed completely. Do not break one text into grammatical units or single words.

Wrong Output: ["THE TREES", "HAVE", "BLOSSOMED"]

Right Output: ["THE TREES HAVE BLOSSOMED"] 

2. Report only what you actually read. Do not guess.
\end{quote}

\end{document}